\documentclass[sigplan,screen]{acmart}

\usepackage{tabularx}
\usepackage{xspace}
\usepackage[ruled, lined, linesnumbered, commentsnumbered]{algorithm2e}
\usepackage{graphicx}
\usepackage{subcaption}
\usepackage{url}

\AtBeginDocument{%
  }

\def\eg{e.g.\@\xspace}
\def\ie{i.e.\@\xspace}

\def\github{\url{https://github.com/ipolyakov/TAGC}}

\copyrightyear{2025}
\acmYear{2025}
\setcopyright{acmlicensed}\acmConference[EuroMLSys '25]{The 5th Workshop
on Machine Learning and Systems }{March 30-April 3 2025}{Rotterdam,
Netherlands}
\acmBooktitle{The 5th Workshop on Machine Learning and Systems (EuroMLSys
'25), March 30-April 3 2025, Rotterdam, Netherlands}
\acmDOI{10.1145/3721146.3721946}
\acmISBN{979-8-4007-1538-9/2025/03}

\begin{document}

\title{TAGC: Optimizing Gradient Communication in Distributed Transformer Training}

\author{Igor Polyakov}
\email{iv.polyakov@vk.team}
\affiliation{%
  \institution{VK, ITMO University}
  \country{Russia}
  }

\author{Alexey Dukhanov}
\email{dukhanov@itmo.ru}
\affiliation{%
  \institution{ITMO University}
  \country{Russia}
  }

\author{Egor Spirin}
\email{egor.spirin@vk.team}
\affiliation{%
  \institution{VK Lab}
  \country{Russia}
}


\begin{abstract}
The increasing complexity of large language models (LLMs) necessitates efficient training strategies to mitigate the high computational costs associated with distributed training.
A significant bottleneck in this process is gradient synchronization across multiple GPUs, particularly in the zero-redundancy parallelism mode.
In this paper, we introduce Transformer-Aware Gradient Compression (TAGC), an optimized gradient compression algorithm designed specifically for transformer-based models.
TAGC extends the lossless homomorphic compression method by adapting it for sharded models and incorporating transformer-specific optimizations, such as layer-selective compression and dynamic sparsification.
Our experimental results demonstrate that TAGC accelerates training by up to 15\% compared to the standard Fully Sharded Data Parallel (FSDP) approach, with minimal impact on model quality.
We integrate TAGC into the PyTorch FSDP framework, the implementation is publicly available at \github.
\end{abstract}





\maketitle

\section{Introduction}
With the rapid advancement of large language models (LLMs), their applications have expanded across various domains, including natural language processing (NLP) and computer vision (CV) \cite{wang2024visionllm}, audio processing \cite{ghosal2023text}, and healthcare \cite{kim2024health}. However, training a state-of-the-art LLMs requires immense computational resources \cite{zhao2023pytorch, rasley2020deepspeed, shoeybi2019megatron}. The trend toward increasingly larger model sizes is evident in models such as DeepSeek \cite{liu2024deepseek}, LLaMA \cite{dubey2024llama}, and GPT \cite{brown2020language}.
This growth in model complexity significantly impacts computational costs, particularly in terms of training time, which can span several weeks or even months. Furthermore, incremental improvements in large models are driven by extensive experimentation, making the execution time of such experiments a critical factor in accelerating advancements in the field.

Multiple methods have been suggested to optimize the LLM training, allowing to scale models faster by improving GPU utilization.
One possible direction for optimization involves the development of more efficient operators that uses GPU-specific computation, such as FlashAttention \cite{dao2022flashattention}, XFormers \cite{lefaudeux2022xformers}, and LigerKernels \cite{hsu2024liger}.
On the other hand, training large-scale neural networks typically requires hundreds of GPUs.
This is achieved through distributed training strategies, including pipeline~\cite{he2021pipetransformer, narayanan2019pipedream}, tensor~\cite{jia2019beyond, narayanan2021efficient} or zero-redundancy~\cite{rajbhandari2020zero, ren2021zero} parallelism, which breaks the training process into parts that are executed in parallel on different computing units.
The model is broken down into a number of shards, each of them is executed on a separate device with periodic synchronization, primarily of gradients.


The synchronization between devices is a component of an overall training performance, and it can be a bottleneck in large models training because the compute amount grows faster than network bandwidth \cite{Gholami07}.
The existing techniques optimize different stages of synchronization between devices. 
For example, tensor quantization both for forward activations~\cite{choi2018pact} and gradients~\cite{alistarh2017qsgd, du2020high, tang2020communication} or gradient sparsification~\cite{basu2019qsparse, tang2020communication}.
This also includes lossless compression methods~\cite{li2024accelerating} for gradients.

Although there are various communication optimization techniques, it is not always possible to integrate them into existing tools for training large neural networks.
One widely used tool for optimizing distributed training is NCCL~\cite{Nccl25}, a library that provides efficient collective communication operations for GPU clusters and is considered the industry standard.
However, NCCL requires that operations have the distributive property, namely $
f(grad_1)+f(grad_2)=f(grad_1+grad_2)
$ for some operation $f$.
This is a constraint that many quantization and sparsification algorithms do not satisfy.
This way, compression algorithms become a more viable approach to improving computational cluster utilization.

One existing method, lossless homomorphic compression~\cite{li2024accelerating} (LHC), is an efficient cross-device communication method using NCCL that allows training neural networks in Data Parallelism setup.
The main idea is that the gradients of some models are sparse, which allows them to be effectively compressed during synchronization and reduce the load on the network.
Experiments have shown more than 3x speed up for models with $\approx 95\%$ sparsity and only 1.2x speed up for the BERT model that uses the transformer~\cite{vaswani2017attention} architecture and had very dense gradients with $\approx 20\%$ sparsity.
The lack of model sharding support and ignoring specifics of transformer architecture make it unsuitable for use with LLM training pipelines.


In this paper, we propose an approach reducing gradient synchronization time during transformer-based model training in the zero-redundancy parallelism mode.
Our method, Transformer-Aware Gradient Compression (TAGC), is based on LHC using all its efficiency, but with adaptation for sharded models and optimized specifically for the transformer architecture.
We also provide several hyperparameters, making it possible to select a balance between accelerating the training speed and maintaining model quality.

\section{Background}

\subsection{Distributed Training}

Distributed training enables scaling by partitioning computations across multiple devices, such as GPUs or TPUs, allowing for more efficient training of large-scale models.
There are several strategies for distributed training, among which Data Parallelism~\cite{li2020pytorch} and its extension Sharded Data Parallelism~\cite{rajbhandari2020zero, zhao2023pytorch} are commonly used.

In Data Parallelism (DP), the dataset is divided into smaller batches, and each computing device processes a separate batch independently.
Each device holds a complete copy of the model and performs forward and backward passes on its local batch.
After computing local gradients, all devices are synchronized using the All-Reduce operation, resulting in an average gradient on each of them.
At the end of the step, optimizer updates model parameters with the same gradients on each device.

The Sharded Data Parallelism or Zero-Redundancy improves DP by dividing the model, its gradients and optimizer states into a number of shards, and stores each shard on a separate device. 
This makes it possible to train bigger models at the cost of increased communication, as there are multiple synchronisations of shards during each forward and backward step.
Main operations are All-Gather (to collect shards from all devices) and Reduce-Scatter (to distribute the shards back on separate devices).

NVIDIA Collective Communications Library (NCCL)~\cite{Nccl25} is commonly used to optimize multi-GPU communication, providing efficient implementations of collective operations such as Reduce, All-Reduce, All-Gather, and Reduce-Scatter.

\subsection{Lossless Homomorphic Compression}

The lossless homomorphic compression method~\cite{li2024accelerating} is a gradient compression technique with efficient in-network communication.
The method introduces two data structures to transfer the gradient: Index and Count Sketch.
Index tracks the indexes of non-zeros gradient values, and Count Sketch stores those values in a compressed manner.
LHC uses the All-Reduce collective communication operation and integrates into Data Parallelism pipelines, allowing for efficient and lossless gradient synchronization.


\begin{table}
  \caption{Compression level relative to the percentage of sparseness in the LHC algorithm. $\theta\%$ means that the synchronization matrix must contain at least $\theta\%$ of zeros, and can be lossless compressed-decompressed with the corresponding compression ratio.}
  \label{tab:sparsification}
  \begin{tabular}{ccl}
    \toprule
    Compression level & Sparsification ($\theta$), \% \\
    \midrule
    $2$x  & $80\%$    \\
    $4$x  & $90\%$    \\
    $10$x & $98.75\%$ \\
  \bottomrule
\end{tabular}
\end{table}

LHC allows applying different levels of compression for Count Sketch with respect to its sparseness percentage. See Table~\ref{tab:sparsification} for details.
Thus, the sparser the data, the higher the level of compression that can be applied.
To decompress, LHC uses two algorithms: peeling and estimation.
The peeling algorithm provides a substantially better quality, but works effectively only when the gradients are sparse and have 80-95\% of zeros.
The estimation works for any gradient sparsity, but introduces noise, which is generally better to avoid.

Due to limitations in NCCL, which does not support the bitwise OR operation needed to correctly synchronize Index between devices, it is simulated using an addition operation.
There are two options to do so: store a 1-bit Index that is faster, but with a higher amount of collisions, or a 4-bit that is stable, but 4 times bigger in terms of communication amount.  

Direct application of LHC to transformer-based architectures is feasible only with an estimation strategy due to the high density of gradient values~\cite{bambhaniya2024progressive}, which leads to a noticeable degradation in quality. 
Furthermore, modern large language models (LLMs) require sharding during training, necessitating the use of the Reduce, Reduce-Scatter, and All-Gather operators for collective communication, which are not supported by LHC.


\section{Method}

Our method, Transformer-Aware Gradient Compression (TAGC), consists of two parts: integrating lossless homomorphic compression into Zero-Redundancy parallelism, where a model is sharded across devices, and its adaptation to transformer architecture for fast and stable communication.
We also provide technical details of integrating and configuring TAGC into PyTorch FSDP~\cite{zhao2023pytorch}, Zero-Redundancy parallelism implementation in one of the most popular deep learning frameworks. 

\begin{algorithm}[t]
\caption{TAGC Communication Hook}
\label{alg:thlc}
\KwIn{
    shard gradient $g_{shard}$,
    sparsfication threshold $\theta$,
    shard gradient accumulator $acc$,
    shard master device $d$
}
\KwOut{
    total gradient on master device $g$,
    updated shard gradient accumulator $acc_{next}$
}
    $g_{shard} \gets g_{shard} + acc$\;
    $g_{shard}, acc_{next} \gets \text{sparsify}(g, \theta)$\;

    $\text{local\_index} \gets \text{create\_index}(g)$\;
    $\text{index} \gets \text{All-Reduce}(\text{local\_index})$\;

    $\text{local\_count\_sketch} \gets \text{compress}(g_{shard}, \text{index})$\;

    $\text{Reduce}(\text{local\_count\_sketch}, d)$\;

    \tcp{Applying gradient only on Master device}
    \uIf{$\text{LOCAL\_RANK} = d$} {
        $\text{count\_sketch} \gets \text{reduced local count sketches}$\;
        $g \gets \text{peeling\_decompress}(\text{index}, \text{count\_sketch})$\;
        \Return $g, acc_{next}$\;
    }

    \Return None, $acc_{next}$\;





\end{algorithm}

\subsection{Gradient Compression in Sharded Setup}

Our method is based on the LHC~\cite{li2024accelerating} compression method, but with respect to collective communications for sharded models.
The original LHC implementation relies on Data Parallelism pipeline, synchronizing both Index and Count Sketch using All-Reduce at the end of the backward pass.
Meanwhile, TAGC only uses All-Reduce for Index synchronization to determine gradients to reduce, while Count Sketches are synchronized by the Reduce operator, which communicates twice less data.

\begin{figure}[h]
  \centering
  \includegraphics[width=\linewidth]{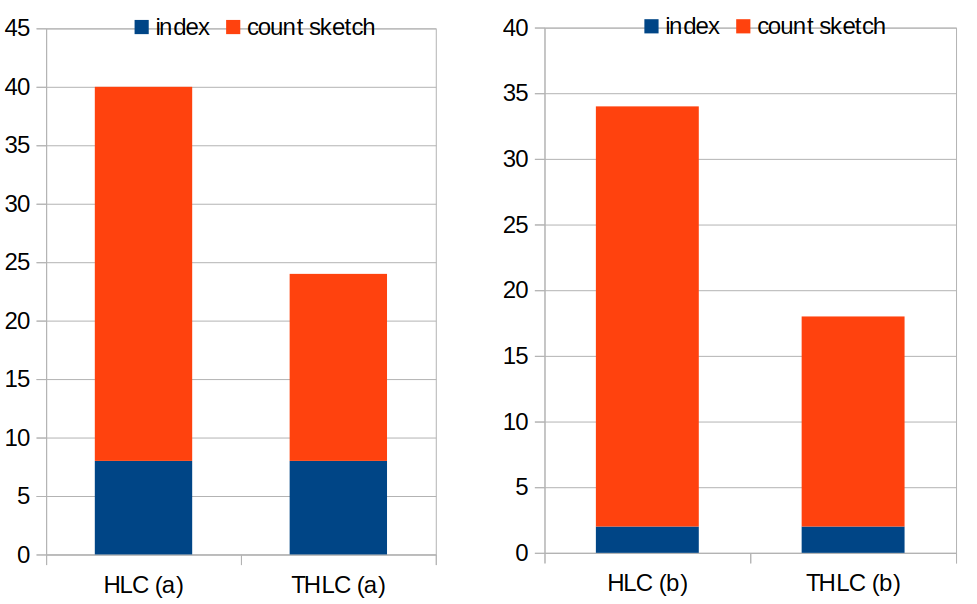}
  \caption{Communication amount per parameter per rank in bits for LHC and TAGC algorithms for 1-bit and 4-bit Index. Count Sketch stores compressed parameters, 2x compression leads to 16-bit per parameter. (a) 4-bit Index for LHC and TAGC algorithms. (b) 1-bit Index for LHC and TAGC algorithms.}
  \label{fig:comms}
  \Description{Examples of the communication amount for the  2x compression ratio for LHC and TAGC}
\end{figure}

\begin{figure}[h]
  \centering
  \includegraphics[width=\linewidth]{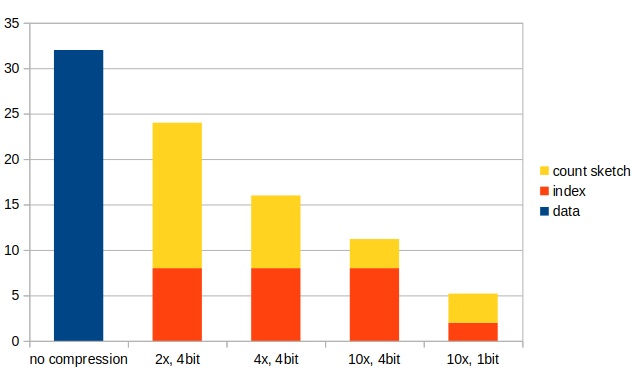}
  \caption{Communication amount per parameter per rank in bits for various compression configurations in TAGC.}
  \label{fig:comm-volume-per-compression}
  \Description{Dependency of communication volume on the compression configuration}
\end{figure}

Figure~\ref{fig:comms} demonstrates the communication amount for the both Index sizes and compressed count sketch with LHC and TAGC algorithms per parameter. Significant improvement in network load can be observed.
Since LHC relies on All-Reduce and TAGC integrates the Reduce operation, our approach provides a 2x advantage in the amount of communication for Count Sketch, that is generally larger than Index per parameter.
Compared to no compression, where each parameter is synchronized as is, TAGC may achieve a 6.15x increase in speed for the 32-bit number representation. Full compression is presented in Figure~\ref{fig:comm-volume-per-compression}.



\begin{figure*}[ht]
    \centering
    \begin{subfigure}{\textwidth}
        \centering
        \includegraphics[width=\textwidth]{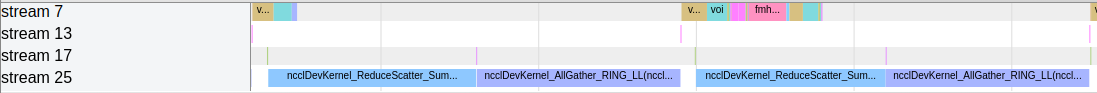}
        \caption{Default FSDP workload without any compression. Computation (streams 7, 13, 17)  takes only 22.4\% of the time. Communication (stream 25) takes 96.1\%. 47.5\% is taken by the gradient exchange only, via Reduce-Scatter.}
        \label{fig:bottleneck}
        \Description{Comparison of the computation and communication time for a backward step when communication is a bottleneck, no compression}
    \end{subfigure}

    \begin{subfigure}{\textwidth}
        \centering
        \includegraphics[width=\textwidth]{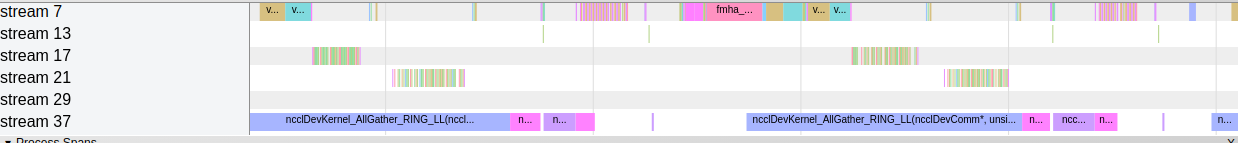}
        \caption{TAGC compression, using 10x compression and 1-bit Index. Computation (streams 7, 13, 17, 21, 29) takes 64.6\% of the time. Communication (stream 37) takes 80.4\%. Gradient exchange takes 20.2\% without gaps, or 42.1\% with gaps.}
        \label{fig:streams}
        \Description{Comparison of the computation and communication time for a backward step after optimizations}
    \end{subfigure}
    \caption{CUDA stream profiles for backward step for 2 layers placed in consecutive FSDP units. Layers are attention projection and feed-forward up-projection. Each row represents a computation or communication stream in an execution timeline. For TAGC, communication time is $32.1\%$ shorter than that of the baseline.}
\end{figure*}

\subsection{Compressing Transformer Gradients}

\paragraph{Gradient Sparsification.}
The main factor to successfully integrate TAGC into training pipeline is the sparseness of gradients. 
Although the transformer gradients are not sparse at all, its gradients follow an approximately log-normal distribution~\cite{chmiel2020neural}.
Based on this, we perform a preliminary sparsification of gradients before their synchronization.
This improvement makes TAGC use a peeling algorithm to decompress and avoid estimation, which leads to a minimal quality drop.

We sparsify each gradient fragment using a sparsity threshold, $\theta$. The value for zeroing the gradient is chosen dynamically, so that at least $\theta\%$ of the values become zero. $\theta$ is a hyperparameter of the algorithm.

Sparsifying gradients which are not zeros means losing information and leads to loss increase. We adapt local gradient accumulation \cite{lin2017deep} to avoid completely losing gradients, which we set to zero. We accumulate the sparsified values of the gradients locally, and we add them to associated gradients on the next iteration.

\paragraph{Selective Compressing.}
Another way to stabilize training is to selectively apply TAGC to model parameters.
This approach allows only large layers with room for sparsity to be compressed and avoid overhead in the compression calculation.
The transformer layer~\cite{vaswani2017attention} consists of two main components: an attention mechanism for transferring information between text tokens, and a feed-forward mechanism, which is applied to each token individually.
There are already numerous optimized attention kernels that fuse operations internally and provide fast computation and communication, \eg FlashAttention~\cite{dao2022flashattention, dao2023flashattention} or Torch SDPA.
Furthermore, there are also normalization layers where parameters are stored, but they are too small to affect the overall communication time.

Therefore, we apply TAGC in two settings: \ie for all layers, and for all non-attention linear layers.
Non-attention linear layers in the transformer model include linear layers in a feed-forward block, an embedding matrix, a language modeling head, and a projection layer after the attention mechanism.

\subsection{Integrating into PyTorch FSDP}
TAGC implementation relies on PyTorch FSDP \cite{zhao2023pytorch}, a Sharded Data Parallelism implementation in one of the most popular frameworks for training neural networks.
The integration is based on various communication hooks implemented in the framework to compress and decompress gradients before and after cross-device communication.

To reduce iteration time, in addition to directly using gradient compression and exchanging the Index and Count Sketch instead of the full gradient, we maximized the overlap of computation and communication operations.
This includes implementation of several callback functions (hooks) within TAGC that would allow parallel execution of operations for the next shard.
For example, when performing a communication step for shard 0, a hook is called that allows computation for shard 1, such as sparsification, and Index creation.

For that, we have replaced the Reduce-Scatter operation in the data exchange stream with a set of compression and compressed data exchange operations in such a way that the data exchange and additional processing operations are performed in parallel where possible.
New and existing operations are synchronized.
The existing operations are performed in four CUDA streams: for performing the backward step, for copying data, auxiliary operations before and after the exchange (to avoid overflow of quantized data types, if they are used), and for communicating data between devices.


In addition, we also overlapped the sparsification and the All-Gather operation of the FSDP framework, which prefetches the parameters of the next block.
The FSDP framework does not support overlapping the All-Gather operation with computation.
This does not cause any problems in the default mode of operation of the FSDP framework, in which gradient accumulation is performed using Reduce-Scatter, since no computation is performed in this case.
When adding the computation of our method, namely sparsification, Index computation, peeling, and so on, it was initially impossible to achieve parallelism between the computation and the All-Gather prefetching.
The main problem is that FSDP places the memory copy operation in the CUDA communication stream, completing the backward step.
We solved this by placing the memory copy operation in the main computation stream.
This allowed us to parallelize the sparsification, initialization, and Index creation with prefetching the next block of parameters.

Overall, TAGC allow to efficiently synchronize gradients for transformer-based model in sharded setup.
This achieved by selective sparsification of gradients, proper selection of collective communication operations, and overlapping communication and computation in the target distributed learning framework.
See Algorithm~\ref{alg:thlc} with pseudocode for more details. Note that it does not include overlapping, since it heavily relies on PyTorch design.

\section{Experiment}

To demonstrate the effectiveness of TAGC, we provide multiple experiments with training a GPT-like~\cite{brown2020language} model on the language modeling problem in a distributed setup.


\subsection{Experimental Setup}

To show the impact of compressing gradients over synchronizing them as they are, we use a setup where communications become a bottleneck.
We use 2 hosts, each of which is equipped with a single NVIDIA GeForce 3080 GPU and connected with a 10 Gbit/sec network.
For the model configuration, we utilize the GPT-2~\cite{brown2020language} architecture and use a small-size configuration with approximately 162 million parameters.
All models in experiments are trained on the OpenWebText~\cite{Gokaslan2019OpenWeb} dataset.

For selective compression, non-attention linear layers represent approximately $82\%$ of all model parameters and, therefore, gradients.


\subsection{TAGC Effectiveness}

The efficiency of TAGC depends on speeding up training without losing quality.
First, we profile the backward step for both the default FSDP implementation and compression with TAGC.
Figure~\ref{fig:bottleneck} provides information for the backward step without compression, in which the computation takes only $22.4\%$ of the time.
In addition, communication takes $96.1\%$ of the time, $47.5\%$ of which is spent exchanging gradients via Reduce-Scatter.
After applying TAGC, as shown in the Figure~\ref{fig:streams}, the computation takes a higher percentage of total time due to the lower communication time.
TAGC spent $32.1\%$ less time for communication than default FSDP.






To demonstrate the effect of reduced communication time, we measured average iteration time during training models. See Table~\ref{tab:iteration} for the results.
Note that applying TAGC for all layers leads to increased iteration time. This is due to the large number of small layers, such as normalizations, which are faster to transmit as is, rather than compress, transmit and decompress.
But when selectively compressing only large linear layers, the iteration time was significantly reduced.

\begin{table}[h]
  \caption{
Average iteration time of the default FSDP and TAGC for both selective compression options. TAGC uses 10x compression with 1-bit Index. FSDP with Non-Attention Linear layers compression stands for the wrapping strategy (whether small layers are stored in the default unit).
}
  \label{tab:iteration}
  \begin{tabular}{cccc}
    \toprule
Algorithm & 
\multicolumn{1}{p{.2\linewidth}}{\centering Compressed \\ Layers} &
\multicolumn{1}{p{.15\linewidth}}{\centering Iteration \\time, sec} &
\multicolumn{1}{p{.2\linewidth}}{\centering Gradient \\ exchange, ms/MB} \\
    \midrule
FSDP  & All            & 26.8 & 0.73   \\
TAGC & All            & 28.5 & 0.64   \\
FSDP & Non-attn  & 25.2 & 0.76 \\
TAGC & Non-attn  & \textbf{22.6} & \textbf{0.51} \\
  \bottomrule
\end{tabular}
\end{table}

\begin{figure}[H]
  \centering
  \includegraphics[width=\linewidth]{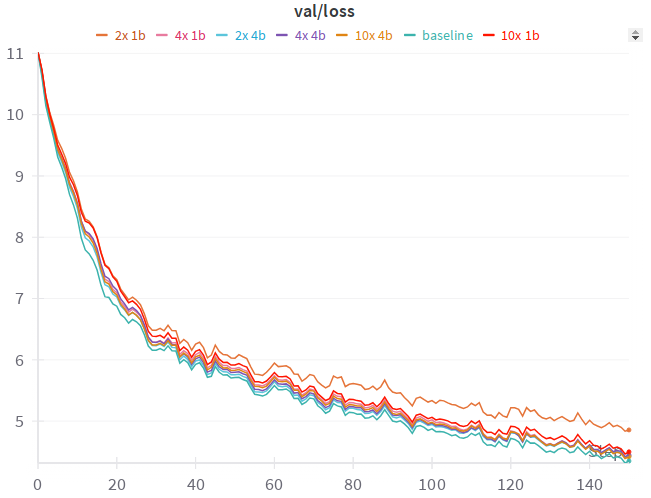}
  \caption{
Validation loss by iteration number for various TAGC configurations and plain FSDP baseline.
}
  \label{fig:loss}
  \Description{Application of compression methods leads to a minor loss increase}
\end{figure}

To estimate quality drop, we trained multiple models in different TAGC setups. Figure~\ref{fig:loss} provides the validation curves during training.
Application of TAGC leads to a minor loss increase.
Difference between default FSDP training and TAGC in the most compressed setup, 10x compression and 1-bit Index, is $3.6\%$.
TAGC with 2x compression and 1-bit index produces an unexpected result with the highest loss value, possibly due to the high number of collisions because of dense gradients together with the unstable index mode.

TAGC reduced the communication time during gradient synchronization under low network bandwidth conditions by more than $30\%$ resulting in a total communication  speed up (together with forward step) of 15\%, with a quality drop of only $3.6\%$.
Moreover, TAGC has several hyperparameters, such as Index size and sparsity threshold, to balance between increasing the speed of communication and the loss in quality.

\section{Discussion and Future Work}

TAGC is a possible optimization algorithm for computational clusters where a network or bus is a bottleneck.
By compressing gradients, TAGC reduces the amount of data transferred at the cost of additional computation for compression and subsequent decompression.
Since TAGC is based on LHC, it inherits the potential of lossless compression, resulting in negligible loss in quality.

Modern computing clusters contain hundreds of GPUs, the main approach to their arrangement is individual machines with 8 devices, which are then connected to a common network through a switch.
Devices per machine are connected with very high bandwidth, intra-node communication and, therefore, TAGC will only slow down the training.
TAGC is useful in inter-node communication, \eg connecting 8 devices from one machine with 8 devices from others, especially when cluster is built without a specialised network, \eg InfiniBand.
One possible direction of future work includes adaption of a hybrid strategy where compression applies only for inter-node communication.
Applying compression for other distributed strategies, \eg Pipeline Parallelism or solely Tensor Parallelism, is also an important direction to enhance cluster utilization.

Another option to improve the method is to combine the gradient compression with quantization.
The current LHC implementation provides solely float32 operations.
A promising direction is to combine LHC compression together with quantization of gradients to float16, bfloat16, or int8 during communication to achieve even better compression.

\section{Conclusion}

In this paper, we introduced the TAGC algorithm for compressing gradients to accelerate communications during distributed training of neural networks.
TAGC is specifically designed to work with the Sharded Data Parallelism training strategy, which is currently one of the most popular ways of training large neural networks.
We also take into account the specifics of transformer architecture, applying layer selective compression with further dynamic sparsification.

As a result, TAGC speeds up training up to $15\%$ compared to default FSDP in low-network bandwidth, with a loss degradation of only $3.6\%$.
This is achieved with the maximum level of compression, where $98.75\%$ of all gradients are sparsified.
TAGC allows selecting the compression level to balance between speed and quality during training.

We implemented TAGC as a part of PyTorch FSDP framework, making it easy to integrate the algorithm in existing pipelines.
To facilitate further research, we made our code publicly available at \github.


\bibliographystyle{ACM-Reference-Format}
\bibliography{sample-base}


\end{document}